# Molecular Strings for LLM Reasoning: Unit–Query Alignment and Representational Locality as Design Principles

Gunwook Nam[1†], Gaeun Sun[1†], and Yousung Jung[1,2*]

1. Department of Chemical and Biological Engineering, Seoul National University, Republic of Korea

2. Institute of Engineering Research, Seoul National University, Republic of Korea

† These authors contributed equally

*Correspondence: yousung.jung@snu.ac.kr

## Abstract

Molecular strings interface chemical structures with large language models (LLMs), yet the features supporting molecular reasoning remain unclear. We propose two design principles: unit–query alignment, whereby units correspond to queried chemical concepts, and representational locality, whereby structural edits remain local in the string. We instantiate both in LAUGHS, a hierarchy of named moieties with explicit linkers, locants and unit boundaries. In instruction-guided editing, LAUGHS raised exact-match accuracy from 48.4% for SMILES to 64.8%, reaching 75.0% with a locality-preserving variant. Introducing query-aligned units improved substructure recognition while reducing generated tokens by up to 70%. LAUGHS also strengthened property-guided redesign, particularly for complex molecules, and produced more high-scoring trajectories on the harder objectives in oracle-guided optimization. Controlled changes to serialization and naming probe the respective contributions of locality and alignment, supporting them as testable criteria for molecular representation design with potential relevance to other structured scientific data.

# Introduction

Large language models (LLMs) are increasingly used to interpret and manipulate chemical information through natural-language instructions. Integrated with scientific tools, they support workflows spanning synthesis planning, reaction optimization, and automated experimentation[1–3]. Molecular string representations provide the interface through which chemical structures can be processed alongside natural language. Applications have progressed from molecular property prediction and reaction modeling[4,5] to molecule–text retrieval, captioning and generation[6,7], followed by chemistry-specific instruction datasets and models[8–10]. Recent approaches have demonstrated property-guided molecular generation through instruction tuning[11] and molecule–text translation through in-context learning[12]. Yet, MolLangBench shows that substantial challenges remain in molecular structure recognition, editing and generation from natural-language instructions[13]. These tasks require LLMs not only to process molecular structures, but also to identify and manipulate the specific chemical concepts referenced in a query.

The choice of molecular representation substantially affects LLM performance. Differences have been observed in molecular property prediction[14], reaction-related tasks[15], and even tokenization and internal embeddings[16]. However, commonly used molecular representations were designed for objectives other than natural-language interaction. SMILES encodes molecular graphs through atom-level traversal, but structurally related atoms can be separated by branching and ring-closure syntax[17]. IUPAC nomenclature exposes named chemical substructures, yet local modifications can require extensive renaming when the parent structure or locant numbering changes[18]. When a local molecular edit changes distant parts of its string representation, the model may need to modify information unrelated to the requested transformation. SMILES Pair Encoding derives substructure tokens from data[19], SELFIES emphasizes syntactic validity[20], and Group SELFIES, t-SMILES and SAFE introduce group- or fragment-level units for molecular generation[21–23]. Although these approaches offer different advantages, their representation units need not correspond to the chemical concepts named in natural-language instructions, and localized structural modifications need not remain localized in their serialized forms.

Contemporaneous studies introduced graph-explicit molecular representations for LLMs, including MolJSON and MoleCode[24,25]. MolJSON encodes atoms and bonds in a structured JSON schema, whereas MoleCode supports persistent graph entities, hierarchical subgraphs and local structural operations, with ring- and scaffold-aware variants. Making topology explicit leaves further choices about which chemical concepts are exposed as units and how structural changes are expressed in serialized text.

Here, we formulate and test two design principles for molecular strings: unit–query alignment, the correspondence between explicit units and queried chemical concepts, and representational locality, the extent to which a local molecular edit remains localized in the string. We examine these principles using LAUGHS (LAnguage for molecUlar Group-aware Hierarchical Strings), a hierarchical string representation that connects named chemical moieties through explicit structural markers and locants. By varying query-relevant units, chemical naming and source–target serialization while holding the underlying molecules and tasks fixed within each comparison, we probe the contributions and limitations of these principles. Evaluations in property-guided generation, redesign and iterative oracle-guided optimization further assess the utility of LAUGHS. These experiments provide an empirical basis for designing molecular representations around the information an LLM must recognize and the changes it must generate.

# Results

## Design Principles for an LLM-Molecule Interface

Two design principles guide the design of molecular strings as interfaces for LLMs. Unit–query alignment concerns the correspondence between representational units and the query targets, supporting access to query-relevant information. Representational locality concerns the confinement of a local molecular change to a contiguous region of the string, reflecting the preservation of chemical locality in the text. We developed LAUGHS to examine these principles.

A molecular graph is encoded as a LAUGHS string in three steps. First, moieties are identified by matching predefined SMARTS patterns to graph substructures. Second, these moieties are prioritized mainly by connectivity, with structural criteria resolving ties. The highest-priority moiety becomes the core and the others become substituents. These moieties are organized into a tree rooted at the core. Third, the tree is serialized using depth-first traversal, with substituents ordered primarily by their attachment locants. This traversal supports representational locality by keeping each substituent branch contiguous. In the string, chemical content units denote moieties, structural marker units specify positions and connectivity, and whitespace separates LAUGHS units. Explicitly delimited, named chemical content units support unit–query alignment. Priority tie-breaking rules are detailed in **Supplementary Note 1** and **Supplementary Table 1**.

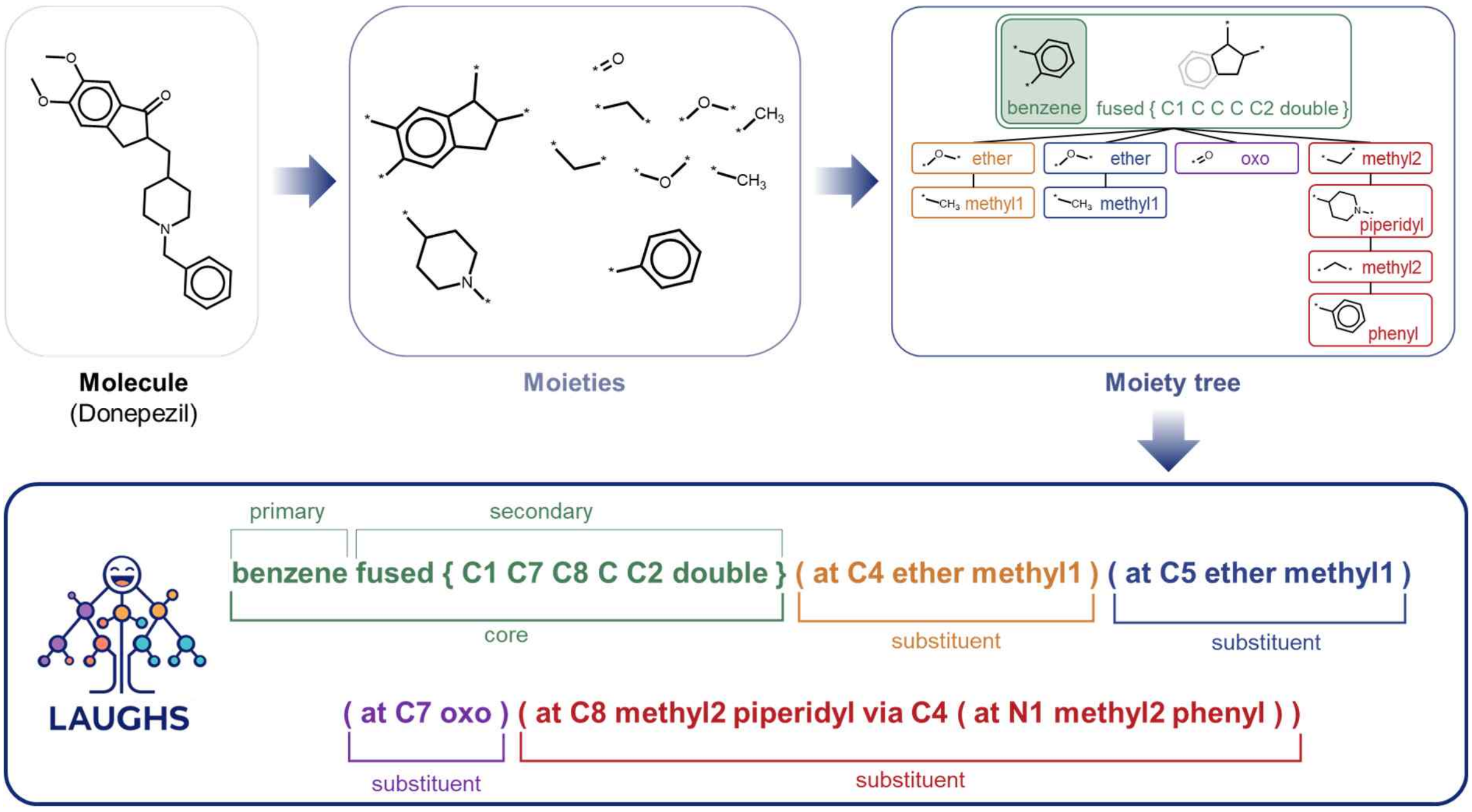

**Fig. 1. Construction of a LAUGHS string from the molecular graph of donepezil.** Identified moieties are organized into a rooted tree, which is then serialized as a string of whitespace-delimited units. Matching colors link nodes of the moiety tree to their corresponding segments in the LAUGHS string, with the core highlighted in green. Labels distinguish the core from its substituents and identify the primary and secondary components within the core. Structural markers include ‘fused’, ‘at’, ‘via’, and locants. Braces enclose the secondary components of the complex ring system, parentheses enclose substituent clauses, and asterisks mark attachment points in the moiety structures.

Complex ring systems are not represented as single named units. Instead, a primary ring is selected based on inter-ring connectivity, and the remaining substructures are described relative to it using identifiers such as ‘fused’, ‘bridged’, and ‘spiro’. **Fig. 1** illustrates the encoding process using donepezil, which contains a fused ring system. Here, chemical content units include ‘benzene’, ‘ether’, and ‘oxo’, while structural marker units include ‘fused’, ‘at’, ‘via’, and locants such as ‘C4’ and ‘N1’.

Tokenization consistency (TC) and semantic similarity (SS) were evaluated for SMILES, IUPAC names, SELFIES, and LAUGHS. The unit-level features set out in **Table 1** are expressed at tokenization and embedding, which TC and SS respectively measure. Under all three tokenizers, TC was highest for LAUGHS and lowest for SMILES (**Supplementary Fig. 1**). Units are delimited by boundaries in SELFIES and LAUGHS, which scored high, but not in SMILES, which scored lowest. SS was highest for LAUGHS and lowest for SELFIES under two of three embedding models (**Supplementary Fig. 2**). SS was high for LAUGHS and IUPAC names, whose units refer to substructures by name. SELFIES explicitly delimits atom-level and structural symbols, showing that boundaries alone do not supply meaning. Functional-group frequencies and mean token counts for the 10,000-molecule evaluation set are reported in **Supplementary Fig. 3** and **Supplementary Table 2**, respectively. The following sections examine how these features affect performance across tasks that edit, describe and generate molecules and how the two principles relate to LLM performance.

**Table 1. Selected unit-level characteristics of English descriptions of molecular structure and molecular representations.** Chemical content describes which molecular components are directly represented, boundaries how units are separated in the string, relations how connections between units are expressed, and organization how units are arranged. English serves as a reference for the linguistic conventions learned during large language-model pretraining. Unit definitions for each representation are given in Methods.

| Representation | Chemical content | Boundaries | Relations | Organization |
|---|---|---|---|---|
| English description | Named substructures | Spaces | Prepositions and verbs | Phrase hierarchy |
| IUPAC name | Named substructures | No uniform delimiters | Locants and affixes | Parent-substituent hierarchy |
| SMILES | Atoms | No uniform delimiters | Bond symbols and ring labels | Graph traversal |
| SELFIES | Atoms | Brackets | Bond prefixes and ring symbols | Graph traversal |
| LAUGHS (ours) | Atoms and named substructures (moieties) | Spaces | Structural markers | Moiety hierarchy |

## Molecular Substructure Reasoning of LLMs

Using the MolLangBench Recognition test set[13] (20 tasks, 3,229 queries; **Supplementary Note 2**) with GPT-5.6 Luna, we examined whether substructure recognition depends on unit–query alignment. Each representation–task pair was assigned one of three alignment levels. Explicit pairs have a unit naming the queried substructure, and patterned pairs a recurring combination of units encoding it. In inconsistent pairs, the substructure is written in context-dependent units or implied by relations among units. Overall accuracy was similar across representations except SELFIES, but task accuracy depended on the alignment (**Fig. 2a** and **Supplementary Fig. 4**). In chiral stereochemistry, LAUGHS and IUPAC names state stereodescriptors directly and exceeded 90%, whereas models using SMILES and its derivatives must infer them and fell below 50%. The gap narrowed in bond stereochemistry, where LAUGHS still scored highest at 88%. Among patterned tasks, α-amino-acid counting favored LAUGHS, which writes this substructure as 'amino1' and 'carboxylic' on one carbon. Penam and steroid counting approached ceiling in all representations except SELFIES. Inconsistent tasks worked against LAUGHS, which writes a benzene ring as 'benzene', 'phenyl', or atoms in braces and has no unit for a quaternary carbon. It fell below 75% on both, whereas the best representations exceeded 88%. The leading representation thus shifted with unit–query alignment. Accuracies across reasoning settings for chiral stereochemistry, α-amino-acid counting and benzene-ring counting are reported in **Supplementary Table 3**.

An aligned unit leaves the model less to derive, so we next examined whether it also reduces generated tokens (**Fig. 2b** and **Supplementary Table 4**). In chiral stereochemistry at high reasoning, representations that must infer stereodescriptors required 5–10 times as many tokens as LAUGHS. LAUGHS led α-amino-acid counting but paid with far more tokens than SMILES and IUPAC names, suggesting the model still had to combine two units. Benzene-ring counting cost LAUGHS about as many tokens as the other representations except SELFIES. At high reasoning, SELFIES used the most tokens on all three tasks but remained among the least accurate representations. Mean numbers of generated tokens per question for all 20 tasks at medium reasoning effort are shown in **Supplementary Fig. 5**.

To test whether inconsistent pairs fail for lack of an aligned unit, we made them explicit by introducing query-relevant units into LAUGHS and compared this with explaining how the queried substructure appears in the prompt (**Fig. 2c**). The four experimental conditions are summarized in **Supplementary Table 5**. At high reasoning, replacing the context-specific units with a single 'benzene' unit raised accuracy from 86.0% to 95.0% and cut tokens by 70%. The '_quaternary' unit raised accuracy from 80.0% to 88.0% while reducing tokens by only 10%, and adding its definition to the prompt reached 97.5% with 70% fewer tokens. Prompt clarification alone cut tokens in both tasks but reached at most 82.5% accuracy. Representing each queried substructure as an interpretable unit thus raised accuracy and cut tokens.

Bond stereochemistry shows that explicit units do not necessarily ensure correct interpretation by the model (**Fig. 2d**). Simply counting the '_E' and '_Z' units of LAUGHS reproduced 99.5% of the benchmark labels across 200 questions. Under the original prompt, however, accuracy ranged from 79.5% to 93.0% across reasoning settings. Most errors were overcounts (108 of 110 over four settings), suggesting that the model also counted double-bond units lacking '_E' or '_Z'. Adding two simple sentences clarifying that only these units denote double-bond stereochemistry raised accuracy to at least 95.5% across all settings, close to the counting reference.

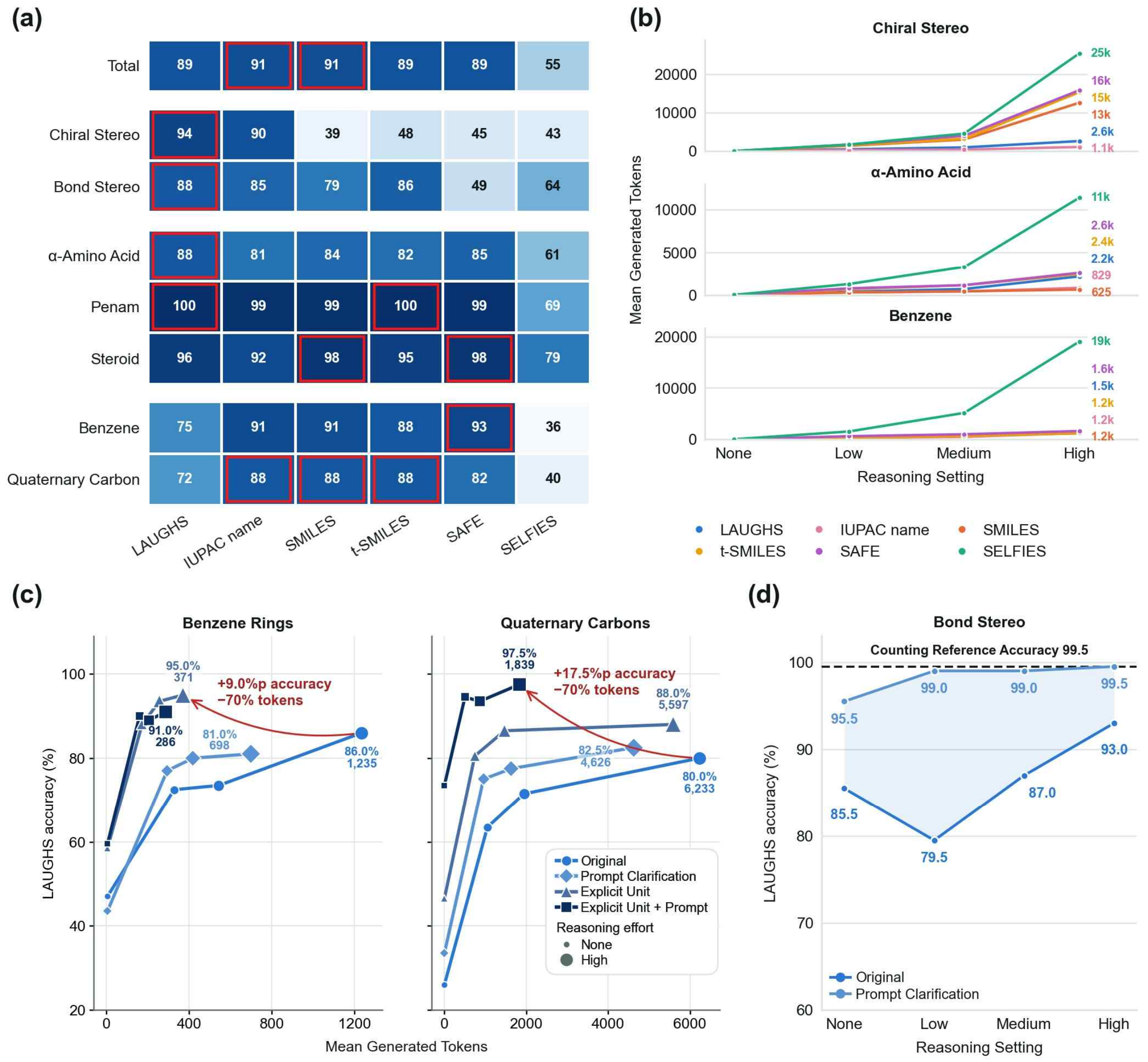

**Fig. 2. Molecular recognition across representations and reasoning settings.** (a) Exact-match accuracy (%) across six representations at medium reasoning. Tasks are separated into blocks by their alignment level in LAUGHS, and red outlines mark row maxima. (b) Mean generated tokens for each query under none, low, medium, and high reasoning settings. Labels at the end of each line represent the number of tokens under high reasoning. k denotes 1,000 tokens. (c) LAUGHS accuracy and mean generated tokens as reasoning effort rises in benzene-ring and quaternary-carbon counting under the original string and prompt (Original), a rule for the queried substructure added to the prompt (Prompt Clarification), a ‘benzene’ or ‘_quaternary’ unit inserted into the string (Explicit Unit), and the modified string accompanied by an explanation of the inserted units (Explicit Unit + Prompt). Marker size increases with reasoning effort from none to high along each line. Labels report accuracy and mean generated tokens at high reasoning effort. At high reasoning, arrows run from Original to the highest-scoring condition, toward higher accuracy and fewer tokens. (d) LAUGHS accuracy on bond-stereochemistry tasks across reasoning settings before and after clarifying ‘_E’ and ‘_Z’ in the prompt. Dashed lines indicate the counting reference. Panels (a) and (b) draw on a common set of 3,229 questions available in all six representations, whereas panels (c) and (d) use 200 questions of each task.

## Instruction-Guided Molecular Structure Editing

To assess how molecular representation affects accuracy in instruction-guided molecular editing, seven representations were compared: IUPAC names[18], SMILES[17], R-SMILES[26], SAFE[23], SELFIES[20], Group SELFIES[21], and LAUGHS. LAUGHS-local, a target-serialization variant that strengthens locality by preserving unchanged input substrings at the cost of canonical unit ordering, was also compared with canonical LAUGHS to isolate the contribution of locality. The instruction dataset comprises four substituent-level and three ring-level editing tasks and was split into training and test sets. The substituent-level tasks drew on 42 named groups (**Supplementary Table 6**). Each representation was used to encode the inputs and targets and to fine-tune an LLM, and all models were evaluated on the same test set. Methods and **Supplementary Note 3** give fine-tuning details and prompt templates, respectively.

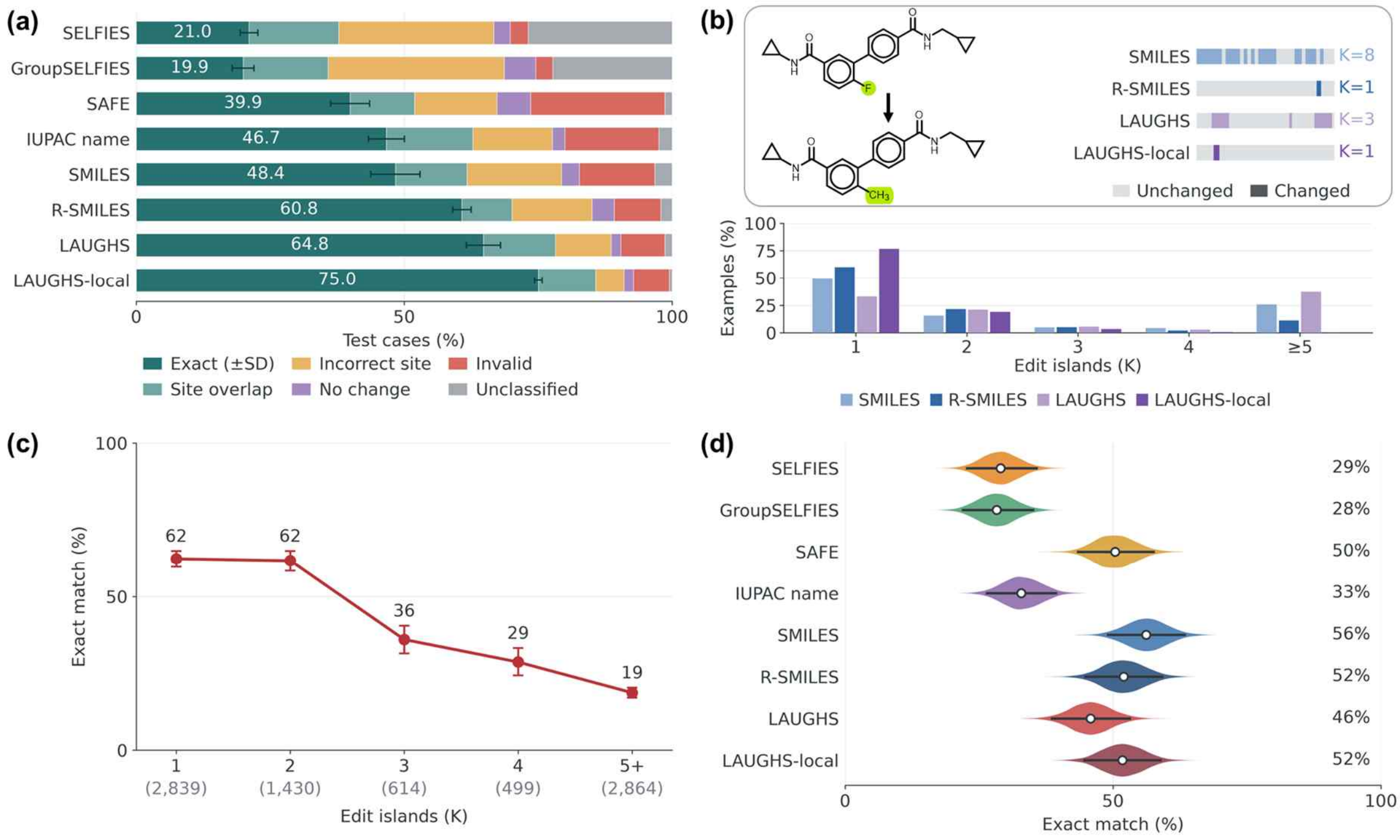

**Fig. 3. Instruction-guided molecular editing across molecular representations.** (a) Outputs on the test set for each molecular representation, classified as exact matches (exact), site overlap, incorrect site, no change, invalid and unclassified. Segment widths and labels show means across the three replicates, and error bars show standard deviations of exact-match rates across replicates. (b) Illustrative substitution of a fluoro group with a methyl group and corresponding token changes across four representations. Changed atoms are highlighted in light green. Each contiguous changed region represents an edit island, and K denotes the number of islands. The corresponding distribution plots show K for each representation over all test examples, aggregated across editing tasks and the three replicates. (c) Exact-match rate as a function of K on the test set. Points show exact-match rates pooled across seven representations, tasks and replicates. Numbers above points show rounded exact-match rates, and numbers in parentheses indicate the number of outputs in each bin. Error bars denote 95% confidence intervals from 5,000 bootstrap resamples of input molecules within each replicate. (d) Ring-atom replacement accuracy across molecular representations. Violin plots show distributions of exact-match rates from 5,000 bootstrap resamples. White circles and percentage labels indicate observed rates averaged across replicates, and black lines denote 95% bootstrap confidence intervals.

Each output was converted to isomeric canonical SMILES where possible and classified into one of six categories (**Fig. 3a** and **Supplementary Table 7**). LAUGHS achieved the highest exact-match rate (64.8%), and LAUGHS-local further increased this to 75.0%. R-SMILES likewise outperformed SMILES (60.8% and 48.4%), differing from it only in atom ordering. SELFIES and Group SELFIES produced few invalid outputs yet reached only 21.0% and 19.9%, their most frequent failure being incorrect-site edits (approximately 30%). SAFE, despite its fragment-based organization, produced more invalid outputs and a lower rate than SMILES. IUPAC names achieved a rate comparable to that of SMILES but with fewer incorrect-site edits and more invalid outputs. These results suggest that validity and fragment-based organization alone do not explain editing accuracy. LAUGHS also outperformed SMILES and IUPAC names for most commercial LLMs under few-shot prompting (**Supplementary Fig. 6**).

The dispersion of changed tokens varied across representations (**Fig. 3b**) and was quantified by the edit-island count (K), the number of contiguous blocks of changed tokens between the input and the target. Across all tasks, single-island edits were more frequent with R-SMILES than SMILES (60.0% and 49.3%) and with LAUGHS-local than LAUGHS (76.7% and 33.1%, **Supplementary Table 8**), with higher accuracy accompanying more localized target changes. Canonical LAUGHS had a lower single-island fraction but higher accuracy than SMILES, indicating that locality alone does not fully explain accuracy differences across representations.

Pooling all 1,178 test examples from the three replicates across seven representations, excluding LAUGHS-local, yielded 8,246 outputs (**Fig. 3c**). Accuracy was similar for one and two edit islands (62.3% and 61.7%) but decreased to 36.0%, 28.7%, and 18.7% for three, four, and five or more islands. Although the decline may not hold within every task or representation (**Supplementary Figs. 7 and 8**), the overall decline is consistent with representational locality affecting editing accuracy.

Across 173 test examples of ring-atom replacement (**Fig. 3d**), SMILES achieved the highest mean exact-match rate (56.2%), followed by R-SMILES, LAUGHS-local, SAFE and LAUGHS (45.8-52.0%), whereas IUPAC names, SELFIES, and Group SELFIES reached only 28.3-32.9%. Replacing a ring atom typically alters a single atom symbol in SMILES but can change the canonical unit ordering of LAUGHS, so fewer LAUGHS targets formed a single edit island (59% and 72%). LAUGHS-local restored this locality with 87% single-island targets, which were edited correctly far more often than multi-island targets (58.7% and 13.0%), yet 62 of the 82 failures were single-island, with some islands spanning 12 to 16 changed tokens. These results suggest that the benefits of representational locality depend on how well it is preserved for a given editing task.

## Property-guided Molecular Generation and Redesign

Unlike structural editing, property-guided generation requires the model to infer molecular structures that satisfy target properties. To assess the effect of the representation on this task, the same base LLM was trained separately with supervised fine-tuning (SFT) on 50,000 identical instruction-molecule pairs encoded as LAUGHS or SMILES (**Supplementary Fig. 9**). Each model was evaluated on 42 instructions covering physicochemical properties, molecular formula and structural motifs, with 1,000 outputs sampled from each instruction. Success rate was the fraction of outputs that were valid and satisfied the condition. LAUGHS raised the overall success rate from 53.4% to 56.8% (**Fig. 4a**) and led on 37 of the 42 instructions, of which ten test molecular formula, the hardest type for both representations (**Supplementary Fig. 10**). The margin was largest for macrocycles, at 83.8% compared with 69.8%, narrowed for structural alerts and LogP, whose 95% intervals included zero, and reversed for molecular weight (**Supplementary Table 9**). For macrocycles, the difference arose among valid outputs, at 90.2% compared with 76.5%, rather than from validity (**Supplementary Fig. 11**). The gain therefore tracks unit–query alignment, being largest where the condition names a structure that LAUGHS writes as a contiguous block of numbered atoms, even without a dedicated macrocycle unit, and smaller for properties of the whole molecule.

Property-guided redesign asks the model to modify a source molecule so that a specified property changes in the requested direction. Each model was trained further with SFT and then direct preference optimization (DPO)[27], which raised both representations by a similar margin. Each model received 200 source molecules and ten natural-language goals to increase or decrease a specified property. The resulting source-goal pairs were written as two instructions, with and without a scaffold-preservation requirement, and ten outputs were sampled from each. Results are reported for the constrained form, since without it both representations replace the source scaffold in nearly every generation (**Supplementary Fig. 12**). LAUGHS succeeded on 75.7% of generations against 63.7% for SMILES, and led in nine of the ten goals, by 7.4 to 25.6 percentage points (%p), with reducing H-bond donors the only exception (**Supplementary Fig. 13**). To test whether the two representations diverge as sources become harder to modify, the 200 sources were stratified into three bins on each of three structural descriptors, the number of aliphatic rings, ring attachments, and the number of stereocenters. The mean difference was larger in each successive bin of every descriptor (**Fig. 4b**). For stereocenters LAUGHS held or improved its success rate as sources became more complex, from 74.4% to 79.3%, whereas SMILES fell from 65.6% to 56.5%, widening the gap from 8.8 to 22.8 %p (**Supplementary Table 10**). This came from structural preservation rather than validity (**Supplementary Fig. 14**), though the highest bin holds 31 of the 200 sources. In the highest bin SMILES produced valid molecules more often, by 4.9 %p, but scaffold retention among valid outputs favored LAUGHS by 26.4 %p. LAUGHS therefore better preserves source structures as molecular complexity increases, consistent with a role for representational locality in property-guided redesign.

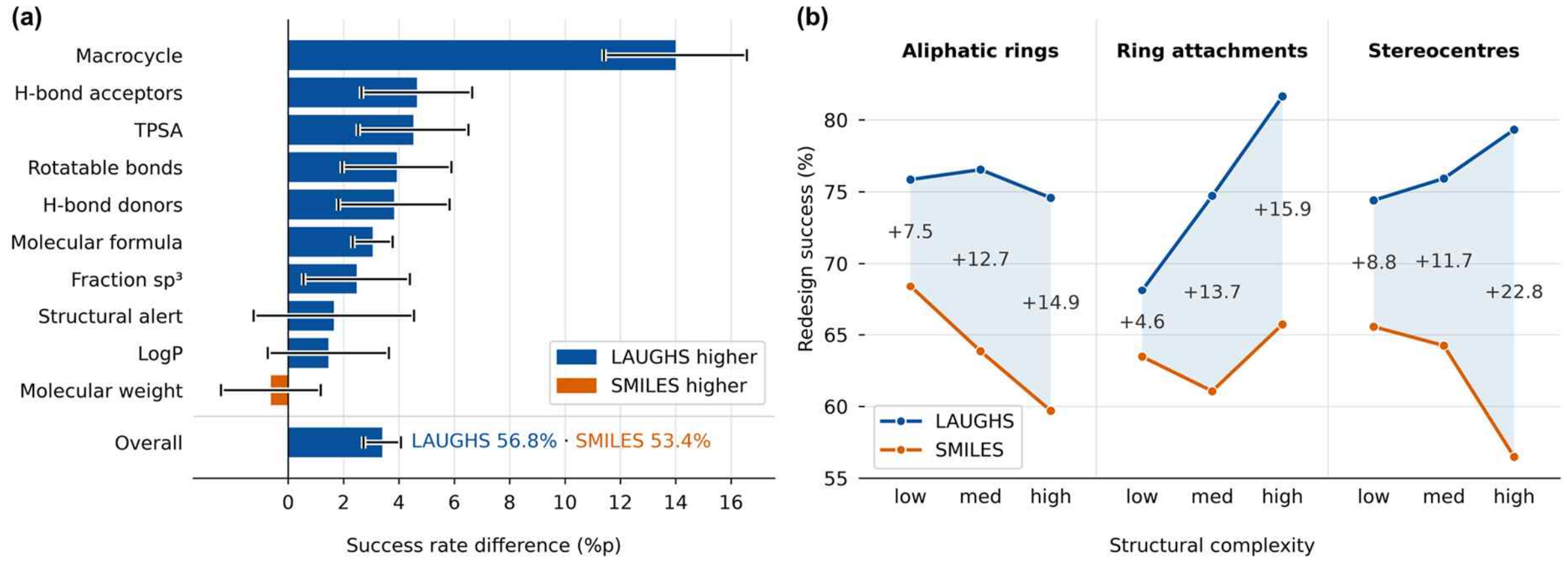

**Fig. 4. Representation effects in property-guided generation and redesign.** (a) Difference in success rate between LAUGHS and SMILES by condition type, for the instruction-tuned LLMs. Each of the 42 single-condition instructions was used to generate 1,000 samples, with the instructions grouped into ten condition types. Bars show LAUGHS minus SMILES in percentage points (%p), with blue indicating higher success for LAUGHS and orange for SMILES. Error bars are 95% Newcombe intervals. The overall row pools all attempts. (b) Redesign success rate for each representation across three source-structure descriptors, for the LLMs trained with SFT and then DPO, under the scaffold-preserving instruction. Lines show the mean success rate across the 200 source molecules, each attempted ten times under each of ten property goals. Labels give the gap between representations (%p). Bin boundaries, the number of sources in each bin and paired 95% bootstrap intervals for these differences are listed in **Supplementary Table 10**.

## Iterative Activity Optimization and Edit Description

To extend molecular redesign beyond a single step, iterative activity optimization proceeds by hill climbing on the oracle score over successive rounds. Three protein-activity objectives were used, DRD2, GSK3β and JNK3, each scored by a TDC activity classifier[28] returning a value between 0 and 1. For SMILES and LAUGHS, training continued from their respective generation SFT adapters, first with SFT on source-target pairs with improved oracle scores, and then with DPO. For each objective, 100 trajectories started from the same source molecules in both representations and were optimized for 50 rounds, with an invalid output leaving the current molecule unchanged. A trajectory counted as high-scoring when its retained molecule reached an oracle score of 0.9 (**Supplementary Fig. 15**). On DRD2 every trajectory reached the threshold under either representation, which separated only on the harder objectives (**Fig. 5a**). LAUGHS led from the early rounds onward on both of these, though SMILES closed the gap on GSK3β in the final rounds, whereas the lead held on JNK3 (**Supplementary Fig. 16**). Cumulative validity remained high for both representations, with LAUGHS higher on GSK3β and JNK3 (**Supplementary Fig. 17**). The advantage of LAUGHS across successive rounds is consistent with representational locality facilitating iterative molecular optimization.

To probe the contributions of locality and chemical naming to edit description, we varied branch ordering and naming in LAUGHS while keeping the molecular transformations fixed (**Fig. 5b**). The four resulting variants and an aligned-SMILES control, which provides locality without chemical naming, were evaluated on the same 102 edit pairs with description generation and judging held fixed. Independently shuffling branches in the source and edited strings reduced LAUGHS accuracy by 14.7 %p, whereas applying the same noncanonical order to both partly recovered the loss. Replacing 17 chemical content units with equivalent SMILES fragments while retaining branch structure caused a smaller decline (4.9 %p), which increased when combined with shuffling. In SMILES, aligning atom traversal improved accuracy by only 3.9 %p, leaving an 11.8 %p gap to LAUGHS. These results support contributions from locality and chemically meaningful units, while the limited gain from aligned SMILES shows that locality alone is insufficient.

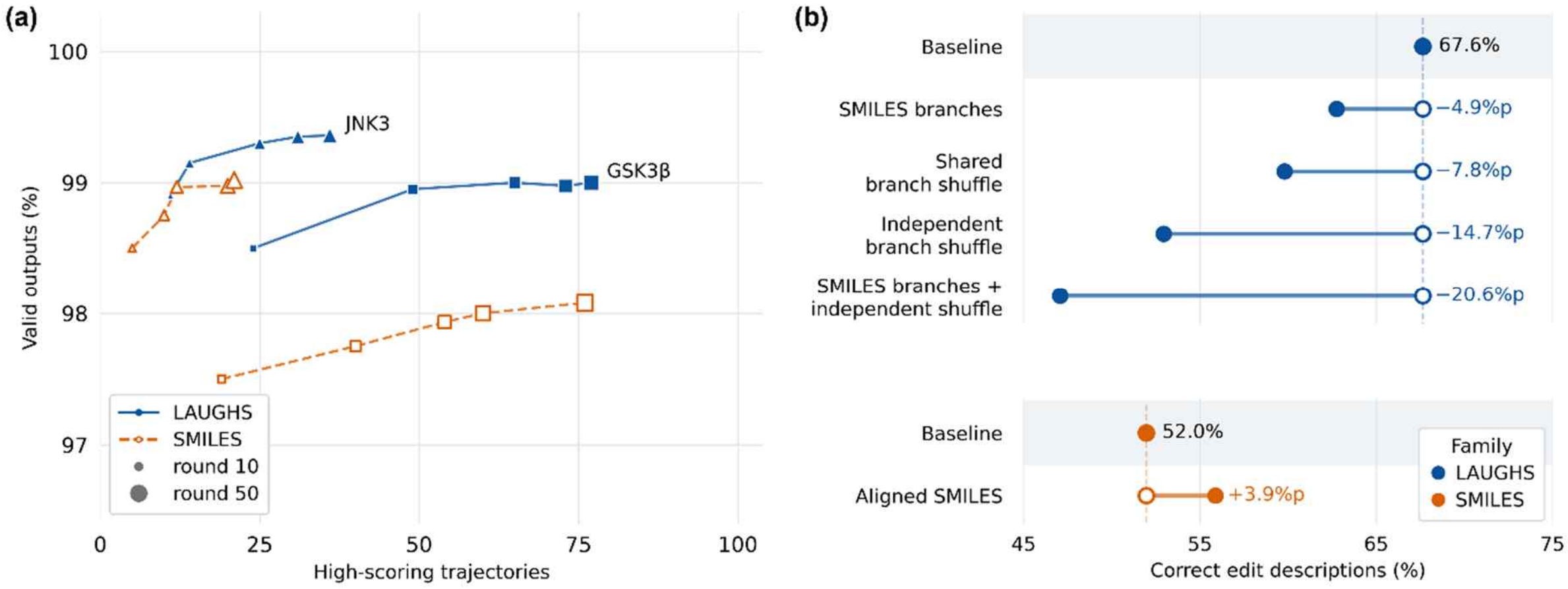

**Fig. 5. Iterative activity optimization and edit description.** (a) Cumulative valid-output rates against high-scoring trajectories for the activity DPO models, run for 50 oracle-guided rounds from 100 source molecules for each objective. A trajectory becomes high-scoring once its retained molecule reaches an oracle score of 0.9, and valid outputs are the share of all attempts so far that parsed. Marker size grows from round 10 to round 50, triangles and squares denote JNK3 and GSK3β, and blue and orange denote LAUGHS and SMILES. (b) Correct edit descriptions for seven representation conditions over the same 102 molecular-edit pairs, generated by GPT-5.4 Mini and graded by GPT-6.1 Sol over three repeats. Shaded rows show each representation's baseline. For other rows, filled circles indicate variants and open circles their respective baselines, with differences labeled in percentage points (%p).

## Discussion

We propose unit–query alignment and representational locality as distinct design principles for molecular representations intended for LLMs. LAUGHS instantiates these principles through named chemical moieties and a hierarchical syntax, while the locality of individual edits also depends on serialization. Building on prior studies demonstrating the utility of explicitly conveying molecular structures to LLMs, we examined which representation features contribute to performance through controlled experiments.

Alignment determined how directly a queried substructure could be read. Accuracy was high when the queried substructure matched an explicit unit or a patterned combination, but fell when it was implied only by relations among units or written in context-specific ones. Introducing a 'benzene' unit raised accuracy from 86.0% to 95.0% while reducing generated tokens by 70%, outperforming prompt clarification alone. However, explicit units did not always ensure correct interpretation, as clarifying their meanings further improved accuracy in the quaternary-carbon and bond-stereochemistry tasks. Locality characterized how concentrated structural changes were between input and target strings. The same molecular edit involved different numbers of changed regions across representations, with more localized changes generally associated with higher editing accuracy. Within LAUGHS, serializing targets to preserve unchanged input substrings raised editing accuracy from 64.8% to 75.0%, whereas shuffling branch order reduced edit-description accuracy by 14.7 %p. However, locality alone did not guarantee accurate editing, as failures occurred even in single-island edits. LAUGHS also outperformed SMILES in property-guided generation, particularly for explicit structural conditions, and in scaffold-constrained redesign of complex source molecules. As interfaces for LLMs, molecular representations should therefore expose what the model must read and localize what it must write.

As a new representation, LAUGHS is less established than SMILES or IUPAC names, and its exposure during model pretraining remains unknown. While this may limit its utility in zero-shot settings, our results demonstrate that LAUGHS can be effectively utilized through few-shot prompting in closed-weight models and SFT and DPO in open-weight models. Its units and hierarchical syntax lengthen strings, which may matter when comparing multiple candidate molecules in a single prompt. However, generated-token counts increased substantially with reasoning effort in the molecular-recognition experiments. A fixed vocabulary also cannot name every rare or domain-specific structure. Complex ring systems, for example, are decomposed into a primary ring and attached substructures rather than written under their trivial names. Future work will add a customizable vocabulary through which users define new moieties as units.

Both principles may extend to other structured scientific representations for LLMs. Reaction representations could expose reaction centers, bond changes, leaving groups, or templates as explicit units, with transformations described locally. Crystal representations could name coordination environments, polyhedral motifs, and defects rather than leaving them to be inferred from coordinates and periodic connectivity. Polymer representations could encode repeat units, block sequences, composition, and branching topology directly, with local changes confined to the corresponding segments. Representation design could thus stand beside model development as a lever for scientific AI.

# Methods

## Terminology

We use four terms with fixed meanings throughout this study. A representation unit, or unit, is the smallest part of a molecular string that the representation's own rules define. The units of a LAUGHS string are its whitespace-delimited parts, such as 'benzene', 'at' and 'C1'. A representation's vocabulary is its set of distinct units. A token is an element produced by an LLM tokenizer. Units and tokens segment the same string by different criteria and need not correspond one to one. A unit may be split across several tokens, and a token may span several units. All token counts we report, including the mean number of generated tokens, are counts of tokenizer tokens, not of units. A moiety is a chemically meaningful substructure that LAUGHS is designed to represent explicitly, such as a ring system, a functional group or a carbon chain. A moiety belongs to the molecule rather than to the string. It takes a single unit when one unit fully specifies its structure, or several when the structure, such as its ring topology or stereochemistry, must be described further through other units.

## LAUGHS Representation

LAUGHS represents molecules as phrases built from named moieties, with a core moiety modified by nested substituent clauses. All units are whitespace delimited. A LAUGHS string is constructed from a molecular graph in three stages: fragmentation, tree construction, and linearization.

**Fragmentation.** The first stage decomposes the input molecule into moieties drawn from a predefined set of named rings, functional groups and carbon chains. SMARTS matching identifies candidate moieties, with larger matches selected first and overlapping matches excluded. Remaining non-ring saturated carbon atoms are grouped into chains, prioritizing the longest available chain. Regions that match no moiety are retained for subsequent structural description. LAUGHS preserves ring bonds and decomposes molecules at non-ring moiety boundaries, retaining the order of each bond, either explicitly with 'double' or 'triple' or implicitly through units such as 'oxo'. These boundaries include connections between distinct moieties, between a moiety and an unmatched region, between separate ring systems, and between ring and non-ring atoms. Decomposition can therefore occur across single, double or triple bonds. Multiple bonds within a selected moiety remain internal. For example, a ring carbonyl can be represented as a ring bearing an oxo group, whereas the C=O bond of an acyclic carbonyl or amide is retained within the corresponding functional-group moiety.

**Moiety tree construction.** The resulting moieties are organized into a rooted hierarchy. A priority scheme selects the global root moiety (the core) using its connectivity, size, atomic composition and neighboring moieties. A specialized rule set handles primary-ring selection, the assignment of additional ring components, and locant ordering within complex ring systems. Ring systems retain their internal connectivity and are described using named ring units such as 'benzene' when they match predefined ring patterns, or at the atom level using the 'ring' unit otherwise. In complex ring systems, this rule applies to the primary ring, while one or more additional ring components are represented by atoms listed in braces. The structural marker units 'fused', 'spiro' or 'bridged' specify the relationships among ring components. Handling ring systems in this way is distinct from fragmentation and does not require cutting ring bonds. These rules define a moiety tree with one node per moiety and one edge for each parent-child attachment. Each edge records the information required to realize that relation in the final LAUGHS string, including the attachment atoms on both moieties, their locants and the connection bond order.

**Linearization.** The encoder then traverses the moiety tree depth first and emits the core followed by its substituent clauses ordered primarily by their attachment locants, with defined tie-breaking rules. A moiety can have different unit names in core and substituent positions, for example 'benzene' and 'phenyl'. The markers 'at' and 'via' introduce the locants on the parent and on the child, respectively. These markers can be omitted when the connection is unambiguous under the grammar. Parentheses delimit substituent clauses, and braces enclose the atoms of a ring. All units are whitespace delimited. For moieties with attachment-count variants, a numerical suffix distinguishes the number of external connections. A primary amino group is written amino1 and a tertiary amino group amino3. Stereocenters and double-bond configurations are annotated at the relevant atoms or bonds using '_R'/'_S' and '_E'/'_Z' in stereochemical mode. The decoder reconstructs the molecular graph from the moiety templates, nested attachments, locants, bond orders and ring-topology descriptions.

**Vocabulary.** LAUGHS defines a vocabulary of 350 units (excluding positional units, as they can be dynamically constructed), of which 319 denote the chemical content of a molecule and 31 are structural markers that specify how that content is arranged. Of the chemical-content units, 90 name a single atom and are written as the name of its element or isotope, covering atomic numbers 1 to 88 together with deuterium and tritium, while 229 identify a chemical substructure in a single word. The latter divide into 34 ring systems, 97 carbon chains and 98 functional groups. These 229 units correspond to 161 distinct moieties, because 68 moieties are named differently depending on whether they appear as the core or as a substituent (e.g. 'benzene' as a core but 'phenyl' as a substituent). LAUGHS thus identifies moieties with a large set of chemical-content units and assembles these units with a small set of structural markers.

## Molecular representation conversion and validation

LAUGHS encoding and decoding were implemented in Python with RDKit (version 2026.03.5), which was also used for BRICS fragmentation and output canonicalization in the molecular-editing experiment. For tokenization consistency and semantic similarity, SMILES, SELFIES and LAUGHS representations were prepared from the source SMILES.

For instruction-guided molecular editing, IUPAC names were generated using ChemScript (version 26.0.0.6141) and validated by SMILES-IUPAC name-SMILES round-trip conversion using OPSIN[29]. SAFE strings were generated using SAFE (version 0.1.14) with BRICS fragmentation, and Group SELFIES (version 1.0.0) used 30 useful and 23 essential groups. R-SMILES inputs were canonical SMILES, and each target was rooted at the first atom in the canonical order of the input that was retained in the target, following the original R-SMILES procedure[26]. Model outputs were decoded using representation-specific procedures and converted to canonical SMILES.

LAUGHS round-trip reconstruction was evaluated on all 2,854,815 compounds in ChEMBL 36[30,31] with available molecular structures by comparing decoded and original molecules as canonical SMILES including stereochemistry. After excluding 15 unparsable inputs, 99.996% of the remaining molecules were reconstructed exactly. Details and the full error breakdown are given in **Supplementary Note 1**.

## Tokenization consistency and semantic similarity

We characterized SMILES, SELFIES, IUPAC names, and LAUGHS with two representation-level metrics: tokenization consistency and semantic similarity. Both used the same 10,000 molecules sampled from ChEMBL, 1,000 from each of ten 50-Da molecular weight bins spanning 250 to 750 Da.

**Tokenization consistency.** Tokenization consistency (TC) measures how consistently representation units are tokenized across molecular strings. After tokenizing each string, we defined the tokenization pattern $p$ of each unit $u$ by the tokens containing all or part of it, their order, and how it appears within each token. For example, the two C units within a single token CCO are assigned different patterns. Let $n(u,p)$ denote the number of occurrences of unit $u$ with tokenization pattern $p$ across the dataset. TC is defined as

$$TC = \frac{\sum_u \max_p n(u,p)}{\sum_u \sum_p n(u,p)}.$$

The numerator sums, over all units, the number of occurrences that follow each unit's most frequent tokenization pattern, and the denominator is the total number of unit occurrences. TC takes values from 0 to 1 and equals 1 when every unit is always tokenized in the same way. SMILES units included complete atom symbols, ring labels, and individual bond, branch, separator, and stereochemical markers. SELFIES units included bracketed symbols and molecular separators. IUPAC names were segmented into alphabetic spans, numeric or alphanumeric locants, descriptor letters (R, S, E, Z, H, D, L, C, F), and individual punctuation marks, with descriptor letters matched before general alphabetic spans. LAUGHS units followed the definitions described above.

**Semantic similarity**. Semantic similarity (SS) measures how well the embedding of a molecular representation can be matched to a natural-language description of a functional group. Functional-group presence was determined from the source SMILES using RDKit fragment detectors, and the 12 most frequent groups were used (**Supplementary Fig. 3**). Each group was described by three English sentences naming it, "A molecule containing one or more {group}", "The molecular structure includes {group}", and "A chemical compound characterized by {group}". Each sentence was embedded and L2-normalized, and the three vectors were averaged and normalized again to obtain the group concept vector. Each molecular representation was embedded with the same embedding model under the fixed input template "Molecular information: {representation}", and its cosine similarity to a concept vector was the prediction score for that group. For each group (g), the area under the ROC curve ($AUC_g$) was calculated against the RDKit-derived presence or absence labels. SS was defined as the macro-average across the 12 groups:

$$SS = \frac{1}{12}\sum_{g=1}^{12} AUC_g.$$

## Molecular substructure recognition

We evaluated 20 of the 23 structure recognition tasks in MolLangBench[13], excluding the one-hop, two-hop, and three-hop neighbor-counting tasks because not all representations can specify individual atom positions. We then selected 3,229 questions that could be converted into all six representations: LAUGHS, IUPAC names, SMILES, t-SMILES (TSID, MMPA_DY), SELFIES and SAFE (MMPA). For each question, the prompt included the query molecule in the evaluated representation and five fixed reference examples showing its conversion from SMILES. For the SMILES condition, the five examples contained only canonical SMILES strings. We evaluated GPT-5.6 Luna (OpenAI: gpt-5.6-luna) with four reasoning settings (none, low, medium, and high). We scored predictions by exact match to the ground-truth counts and recorded the mean number of generated tokens, including reasoning and answer tokens.

We evaluated the effects of explicit structural units and prompt clarification on benzene-ring and quaternary-carbon counting tasks, using 200 molecules per task. Four conditions were compared by varying the explicit inclusion of a query-relevant unit in the LAUGHS string and clarification in the prompt. The "original" condition used the unmodified LAUGHS string and prompt. The "prompt clarification" condition added a short paragraph explaining how to identify benzene rings or quaternary carbons, whereas the "explicit unit" condition modified the LAUGHS string to explicitly denote the queried structure ('benzene' or '_quaternary') without changing the prompt. The "explicit unit + prompt" condition combined the modified LAUGHS strings with an explanation of the inserted feature units. Each condition was evaluated with GPT-5.6 Luna at none, low, medium, and high reasoning settings. The four conditions are summarized in **Supplementary Table 5,** and the prompt templates is provided in **Supplementary Note 3**.

For bond-stereochemistry counting, we computed a counting reference accuracy by directly counting the '_E' and '_Z' units in each LAUGHS string and comparing the resulting counts with the benchmark labels. This analysis included all 200 bond-stereochemistry questions. The reference accuracy was 99.5% because, for one complex polycyclic molecule, a double bond shared by two fused rings carries its '_E' or '_Z' unit in both rings, yielding a count one greater than the number of distinct double bonds. We defined the difference between this counting reference accuracy and the observed model accuracy as the recognition gap. To test whether this gap reflected ambiguity in interpreting LAUGHS units, we appended a short prompt clarification stating the LAUGHS unit rule that '_E' and '_Z' indicate double-bond stereochemistry. We evaluated the original and clarified prompts on all 200 questions using GPT-5.6 Luna at none, low, medium, and high reasoning settings. To characterize error patterns, we also compared the number of units denoting double bonds without '_E' or '_Z' in the strings of correctly and incorrectly answered questions under high reasoning.

## Instruction-guided molecular editing

The instruction-guided editing dataset was built from ChEMBL 36. Three independent replicates of the instruction dataset were constructed by random sampling, each with 500 input molecules for each of seven edit types and no input molecule shared across replicates. The substituent-level edits were addition, deletion, replacement and swap of named substituents, and the ring-level edits were ring expansion, ring contraction and ring-atom replacement within ring systems of two to five rings. Source molecules were filtered by heavy-atom count, ring count, and formal charge. Molecules were retained only when the instruction identified a unique edit site and the edit produced a valid target. Each retained molecule was paired with a natural-language instruction and its edited target, giving 3,500 examples in each replicate. Examples were encoded using seven molecular representations (SMILES, SELFIES, IUPAC names, Group SELFIES, SAFE, R-SMILES, and LAUGHS) and an additional LAUGHS-local configuration. LAUGHS-local is a structurally modified target representation of the input LAUGHS that preserves unaffected substructures, atom locant numbers, and relative branch order. Substituents and ring atoms are added at the specified sites, while replacements occur at the existing sites, preserving the locant numbers of retained atoms. Added substituents are written after existing branches, and ring expansion assigns consecutive locants above the maximum in the corresponding scope. Substituent swaps exchange only the branches at the two attachment sites. LAUGHS-local decodes to the same connectivity and stereochemistry as the corresponding canonical LAUGHS and is used only for target serialization, while model inputs remain canonical LAUGHS. Dataset construction details are provided in **Supplementary Note 4**.

For each replicate, examples were divided into training, validation and test sets at an approximate 80:10:10 ratio, based on the Bemis-Murcko scaffolds of the input molecules, with no input scaffold shared across splits. The same example assignments were used across representation configurations within each replicate. Ministral-3-14B-Base-2512[32] was fine-tuned separately on each configuration in each replicate using QLoRA[33] to generate an edited molecule from an instruction and input molecule. QLoRA used 4-bit NF4 quantization. Adapters with a rank of 16, an alpha of 32 and a dropout rate of 0.05 were applied to the model's attention and feed-forward projections. All models were trained for three epochs with a maximum sequence length of 2,048 tokens. Optimization used paged 8-bit AdamW with a learning rate of $2 \times 10^{-5}$, a cosine learning-rate schedule with 3% warmup, an effective batch size of eight, and zero weight decay. Gradients were clipped to a maximum norm of 0.3, and the cross-entropy loss was computed only over target tokens. The model with the lowest validation loss was evaluated with greedy decoding.

Outputs matching the target isomeric canonical SMILES were classified as exact matches, whereas outputs that could not be decoded into valid molecules were classified as invalid. Among valid non-exact outputs, molecules identical to the input were classified as no change. For the remaining outputs, the modified sites were compared with the intended edit sites. Outputs with overlapping edit sites were classified as site overlap, whereas those with no overlap were classified as incorrect-site edits. All remaining valid non-exact outputs were classified as unclassified.

For the pooled edit-island analysis, all 1,178 test examples from the three replicates were included. K was calculated from the input and reference target using the tokenizer of the shared base model. Exact-match rates were pooled within each K bin.

## Property-guided molecular generation

Instructions for property-guided generation were built from ten condition types. Seven were numeric properties, each split into four intervals: molecular weight at 300, 400 and 500, LogP at 3, 4 and 5, TPSA at 60, 90 and 140, H-bond donors at 1, 3 and 5, H-bond acceptors at 3, 5 and 10, rotatable bonds at 3, 7 and 10, and fraction $sp^3$ at 0.2, 0.4 and 0.6. The other three were the presence or absence of a macrocycle, the same for a structural alert, and an exact molecular formula drawn from ten, giving 42 conditions in total. Macrocycles were rings containing at least eight atoms. Structural alerts were defined as matches to PAINS patterns[34] identified using RDKit, matches to 11 undesirable substructure patterns adopted from SmileyLlama[11], and the presence of an atom shared by at least three SSSR rings. The ten formulas were C12H9ClN2O, C18H10N4O, C19H27N3O2, C19H18N4O2S, C21H30O7, C23H22ClN5O, C22H23F3N8, C24H20FN3O4S2, C33H33N3O5, and C33H39Cl2F3N6O3. Each instruction was written from a chosen ChEMBL 36 molecule, stating conditions that it satisfied and taking it as the paired answer. Five percent of instructions stated no condition, and each condition type appeared in exactly half of the rest. Splitting by molecule gave 50,000 instructions for training and 2,632 for validation, identical across the two representations. Two smaller training sets of 10,000 and 20,000 instructions, with 526 and 1,053 validation instructions, were built in the same way to measure the effect of training size (**Supplementary Fig. 9**). Qwen3-14B[35] was fine-tuned with low-rank adaptation (LoRA)[36] under supervised fine-tuning (SFT), using separate adapters for LAUGHS and SMILES. Both adapters were trained for three epochs with a LoRA rank of 16, a LoRA alpha of 32, a dropout rate of 0.05, an effective batch size of 8, a learning rate of $2\times10^{-4}$ and a maximum sequence length of 1,024 tokens. The models were evaluated with one instruction for each condition, each sampled 1,000 times. Sampling used a temperature of 1.0 and allowed up to 1,024 new tokens, with the thinking mode of Qwen3 turned off.

## Property-guided molecular redesign

Training pairs for property-guided redesign were drawn from ChEMBL 36[30,31], each pairing a source molecule with a target molecule that differed in the goal property. The model received the source molecule and a natural language goal, and was trained to produce the target. Molecules were filtered on heavy atom count, connectivity and the presence of a ring, then matched within shared generic scaffold groups. A match was kept when the target moved the goal property in the requested direction by at least a property-specific threshold (**Supplementary Table 11**). Instructions came in two types, one asking for the source scaffold to be kept and one leaving it free. The scaffold-preserving type required identical Bemis-Murcko scaffolds with matching atom and bond types, a Morgan radius 2 Tanimoto similarity to the source of at least 0.5 and a heavy atom ratio between 0.7 and 1.3. The scaffold-free type lowered the similarity floor to 0.2 and widened the ratio range to 0.5 to 1.5. Each combination of goal and instruction type contributed 10,000 pairs, giving 200,000 training pairs, with a further 10,000 pairs for validation and 10,000 for testing. Splits were made by generic scaffold, with the 11,350 scaffold groups represented in the training pairs. The 200 final evaluation source molecules shared neither identical molecules nor generic scaffolds with the redesign SFT training and validation sets. The generation SFT adapters were trained further on these pairs, which are identical across the two representations, continuing the same LoRA configuration, for one epoch with a learning rate of $1\times10^{-4}$, an effective batch size of 8 and a maximum sequence length of 2,048 tokens.

Preference pairs were built from pooled generations of the LAUGHS and SMILES redesign-SFT models. The 400 source molecules were drawn from the SFT validation set and shared no molecules with the training set or final test sources. For each source, ten outputs per model were sampled for each of ten property goals under both instruction types. Both generations in a pair came from the same instruction. The highest-scoring generation, based on property improvement, structural bonuses, and a molecular-size penalty, was paired with up to four failed generations, each from a different failure category. Invalid or disconnected generations were discarded. The resulting preference pairs were split by generic scaffold into 14,279 training and 1,222 validation pairs, with the same molecules and pairs in both representations. Training continued from the redesign SFT adapters for one epoch with the same

LoRA configuration, a β of 0.1, a learning rate of $5\times10^{-6}$, an effective batch size of 8 and a maximum sequence length of 2,048 tokens.

Both redesign models were evaluated on the same 4,000 instructions, built from 200 source molecules in the test split, each paired with ten property goals in both instruction types. Ten outputs were sampled from each instruction under the same decoding settings as the generation task. Success required a valid, connected molecule that differed from the source and moved the target property in the requested direction by at least the specified minimum. Scaffold-preserving instructions additionally required the generated molecule to contain the Bemis-Murcko scaffold of the source with matching atom and bond types, without enforcing stereochemistry, and to meet the similarity and size limits specified above.

## Iterative activity optimization

Three protein-activity objectives were used, DRD2, GSK3β and JNK3, scored by the pretrained molecular-activity classifiers from Therapeutics Data Commons (PyTDC v1.1.15)[28], each returning a value between 0 and 1. Models for this task continued from the generation SFT adapters. Training pairs were drawn from 600,000 ChEMBL molecules with oracle scores, keeping one pair per source and objective when the target improved the oracle score by at least 0.20 and its Morgan radius 2 Tanimoto similarity to the source was between 0.3 and 0.5, without scaffold or size constraints. Sampling from that pool gave 60,000 training and 3,000 validation examples for each representation, divided equally among the three objectives. The training and validation sets shared no molecules across source or target molecules, even after removing stereochemistry. Both representations used identical pairs in the same order, with no exact molecular overlap between either split and the 100 evaluation sources per objective. The generation SFT adapters were trained further on these pairs for three epochs with the same LoRA configuration, an effective batch size of 32, a learning rate of $2\times10^{-4}$ and a maximum sequence length of 1,024 tokens, with early stopping on validation loss.

The activity SFT adapters were then trained further with DPO. Parent molecules were stratified by objective and by current oracle score into four bins, eight proposals were generated for each parent from each representation's SFT model and deduplicated by canonical output, and at most two preference pairs were kept for each parent. A preferred generation had to raise the oracle score by at least 0.05 over its parent. Parents, preferred generations and rejected generations are the same molecules in both representations, and test parent molecules and scaffolds were excluded from both splits. This gave 360 training pairs and 135 validation pairs. Training continued from the activity SFT adapters, against a frozen copy of the same adapters as the reference policy, for up to ten epochs with early stopping on validation loss, a β of 0.1, a learning rate of $2\times10^{-6}$, an effective batch size of 8 and a maximum sequence length of 2,048 tokens. The resulting change in the mean final oracle score is shown in **Supplementary Fig. 18**.

Each activity DPO model was then run for 50 rounds from the same 100 source molecules across all three objectives and both representations. In each round, the model was prompted to redesign the current molecule to increase predicted activity for the specified target. The model proposed one molecule, which replaced the current molecule only when its oracle score was higher. An output that failed to parse cost the round and left the current molecule unchanged, so no trajectory ended early and every round added 100 attempts. Decoding used sampling at a temperature of 1.0 with a limit of 1,024 new tokens and the thinking mode of Qwen3 turned off. Oracle scores were cached by molecule, so a molecule generated again was not scored again, and an output that failed to parse could not be scored and cost no oracle call. Comparisons between the two representations are made at equal numbers of rounds (**Supplementary Fig. 19**).

## Edit description and evaluation

Pairs of a source molecule and its edit were collected from the activity optimization trajectories generated by the LAUGHS and SMILES activity-DPO models. We randomly selected 17 pairs of consecutive valid molecules with localized structural differences from each model for each of the three activity targets, requiring a maximum common substructure covering at least 60% of the larger molecule's heavy atoms and localized structural differences. This yielded 102 molecular pairs, each represented in both LAUGHS and SMILES, giving 204 representation-specific pair instances.

Each pair was presented under seven representation conditions built from the same two molecules. SMILES branches replace the chemical names of LAUGHS branches with atom-level SMILES fragments, covering seventeen names, chloro, fluoro, bromo, iodo, nitro, cyano, hydroxy, methyl1 to methyl4, amino1 to amino3, amide, ether and ethyl. Shared branch shuffle reorders branches using the same order in both molecules, and independent branch shuffle draws a separate order for each. Aligned SMILES writes the two molecules with matched atom ordering.

One description of the structural change was written for each pair and condition by GPT-5.4 Mini (OpenAI: gpt-5.4-mini) without reasoning, and the descriptions were graded by GPT-6.1 Sol (OpenAI: gpt-6.1-sol) at medium reasoning effort over three independent repeats, with a component-wise majority over the edit, the site and the absence of extra claims. A description was judged correct only when all three criteria passed. For every condition the judge was shown the same canonical SMILES and LAUGHS of the two molecules, with only the description changing, and the condition was never named in the prompt. Prompt templates are provided in **Supplementary Note 3**.

Two human reviewers each graded the same 204 baseline descriptions based on chemical structure diagrams of molecules and the same correctness criteria, with representation identities and judge verdicts hidden. Final agreement between reviewers was 96.6% (**Supplementary Table 12**). Averaged across reviewers, agreement with the LLM judge was 86.5%, with a mean Cohen's $\kappa$ of 0.714 and a macro F1 of 0.857 over the correct and wrong classes (**Supplementary Table 13**).